\documentclass[letterpaper, 10 pt, journal, twoside]{ieeetran}  
\usepackage{cite}
\usepackage{hyperref}

\hypersetup{
    colorlinks=true,
    linkcolor=blue,
    filecolor=magenta,      
    urlcolor=blue,
    }

\IEEEoverridecommandlockouts                              

\usepackage{graphics} 
\usepackage{epsfig} 
\usepackage{mathptmx} 
\usepackage{times} 
\usepackage{amsmath} 
\usepackage{amssymb}  
\usepackage{lipsum}

\usepackage{algorithm,algcompatible}
\usepackage{algorithmicx}
\usepackage{algpseudocode}
\usepackage{amsmath}

\usepackage{makecell}

\begin{document}
\title{A Minimalistic Stochastic Dynamics Model of Cluttered Obstacle Traversal}

\author{Bokun Zheng, Qihan Xuan, and Chen Li$^{1}$%

\thanks{Manuscript received September 10, 2021; Revised December 14, 2021; Accepted February 06, 2022.}
\thanks{This paper was recommended for publication by Editor Xinyu Liu upon evaluation of the Associate Editor and Reviewers' comments.
This work was supported by a Burroughs Wellcome Fund Career Award at the Scientific Interface and The Johns Hopkins University Whiting School of Engineering start-up funds to C.L. (Corresponding author: Chen Li.)}
\thanks{$^{1}$The authors are with the Department of Mechanical Engineering, Johns Hopkins University, Baltimore, MD 21218 USA (email:  bzheng8@jhu.edu; qxuan1@jhu.edu; chen.li@jhu.edu). }%
\thanks{Digital Object Identifier (DOI): see top of this page.}

}


\markboth{IEEE Robotics and Automation Letters. Preprint Version. Accepted February 2022}
{Zheng \MakeLowercase{\textit{et al.}}: A minimalistic stochastic dynamics model of cluttered obstacle traversal} 

%

\maketitle

\begin{abstract}

Robots are still poor at traversing cluttered large obstacles required for important applications like search and rescue. By contrast, animals are excellent at doing so, often using direct physical interaction with obstacles rather than avoiding them. Here, towards understanding the dynamics of cluttered obstacle traversal, we developed a minimalistic stochastic dynamics simulation inspired by our recent study of insects traversing grass-like beams. The 2-D model system consists of a forward self-propelled circular locomotor translating on a frictionless level plane with a lateral random force and interacting with two adjacent horizontal beams that form a gate. We found that traversal probability increases monotonically with propulsive force, but first increases then decreases with random force magnitude. For asymmetric beams with different stiffness, traversal is more likely towards the side of the less stiff beam. These observations are in accord with those expected from a potential energy landscape approach. Furthermore, we extended the single gate in a lattice configuration to form a large cluttered obstacle field. A Markov chain Monte Carlo method was applied to predict traversal in the large field, using the input-output probability map obtained from single gate simulations. This method achieved high accuracy in predicting the statistical distribution of the final location of the body within the obstacle field, while saving computation time by a factor of $10^5$ over our dynamic simulation.
\end{abstract}


\section{INTRODUCTION}

\IEEEPARstart{M}{any} important applications require robots to traverse cluttered obstacles such as search and rescue in rubble, environmental monitoring on the forest floor, and extraterrestrial exploration through Martian rocks. The dominant approach of robotic locomotion in complex environments is to avoid obstacles \cite{borenstein1991vector,rimon1990exact,khatib1986real,thrun2010toward}, which is challenged in such cluttered terrain. By contrast, insects such as the discoid cockroach are excellent at traversing cluttered large obstacles, often using their bodies and appendages to make physical interaction with obstacles \cite{harley2009characterization, li2015terradynamically,gart2018body,gart2018dynamic,othayoth2020energy,han2021shape,wang2021cockroaches}. Understanding how the physical interaction between a self-propelled locomotor and obstacles leads to or hinders traversal can inform how to modulate physical interaction to control robot locomotion beyond avoiding obstacles (\cite{li2015terradynamically,gart2018body,gart2018dynamic,othayoth2020energy,han2021shape,mi2021omniroach,xuan2021environmental}; for a review, see \cite{othayoth2021locomotor}) and increase the terrain that they can access \cite{Chen2022the}.

Recent animal and robot studies from our lab established a potential energy landscape approach for modeling locomotor-obstacle interaction, which provided insights into the mechanical principles of how different modes of locomotion emerges from the interaction, which can be controlled to enhance or supress locomotor transitions to achieve obstacle traversal \cite{othayoth2021locomotor,othayoth2020energy}. Our study is inspired by and builds upon our previous work \cite{othayoth2020energy}. The potential energy landscape approach revealed that, when interacting with large obstacles, although the motion of a self-propelled locomotor (animal or robot) is stochastic, it displays highly stereotyped modes \cite{othayoth2021locomotor}. The stereotyped modes results from the locomotor being strongly attracted to distinct basins of a potential energy landscape during highly dissipative large obstacle interaction \cite{othayoth2021locomotor, othayoth2020energy}. To escape from attraction to certain basins that lead to unfavorable modes (e.g., become trapped against the obstacles \cite{han2021shape, othayoth2020energy}) and transition to other basins that results in favorable modes (e.g., go through gaps between obstacles), the locomotor can use both passive kinetic energy fluctuation from cyclic self-propulsion \cite{othayoth2020energy} and active sensing and control \cite{wang2021cockroaches}. However, there remains a lack of understanding of the dynamics of cluttered obstacle traversal, which is not modeled by the potential energy landscape approach.

Here, we take the next step towards this by developing a dynamic model of locomotor-obstacle interaction using a minimalistic model system. We focus on modeling the stochastic dynamics of a feedforward self-propelled locomotor and do not consider active sensing and control. Using the dynamics model developed, we further explore several questions of interest to biology and robotics: How do self-propulsion and random forces, which are common in animal locomotion and beneficial to survival \cite{bergman2000caribou,faisal2008noise,  li2015terradynamically,moore2017unpredictability,xuan2020randomness}, interplay to affect traversal? What is the effect of terrain asymmetry common in nature \cite{grand2010motion}? How does locomotor-obstacle interaction at a small scale lead to traversal over large spatiotemporal scales?

Our minimalistic model system consists of a circular locomotor body moving on a level plane and interacting with a pair of grass-like beam obstacles which rotate within the plane. The system has four degrees of freedom: two translational degrees of freedom for the body and one rotational degree of freedom for each beam. Besides a constant forward self-propulsive force, a lateral random force to introduce body oscillation, leading to stochastic dynamics. We systematically varied propulsive and random forces and asymmetric beam stiffness and studied their impact on traversal. We also studied the effect of asymmetry in the terrain using the two beam obstacles with different torsional stiffness.

Another goal was to use the systematic results from our stochastic dynamics simulation to test how well the potential energy landscape qualitatively predicts traversal dynamics without solving equations of motion. This idea was inspired by the problem of microscopic protein folding \cite{onuchic2004theory}. Analogous to our system, a free energy landscape approach has proven extremely useful in understanding and predicting how proteins stochastically transition through intermediate states to fold without having to solve equations of motion that are often intractable due to the huge number of degrees of freedom of the system (for the plausibility of this analogy, see \cite{othayoth2020energy,othayoth2021locomotor}).

Finally, we extended the single locomotor-obstacle interaction model to simulate traversal of a larger terrain with multiple obstacles in a lattice configuration. Again, inspired by the field of protein folding problem \cite{chodera2006long}, we tested the hypothesis that the longer-time statistical dynamics of large obstacle field traversal can be well described by a discrete-state Markov chain model constructed from smaller-scale trajectories of the body-obstacle interaction system.

We introduce how the model was defined and the dynamics modeled (Sec. \ref{chap2A}, \ref{chap2B}), how the potential energy landscape was calculated (Sec. \ref{chap2C}), and how we generated a multi-obstacle field (\ref{chap2D}). We then describe results using the stochastic dynamics simulation to study traversal of a pair of beams and testing how well the landscape model informed traversal (Sec. \ref{chap3}) and how well the traversal of a multi-obstacle field was predicted by the Markov chain Monte Carlo method (Sec. \ref{chap3C}). Finally, we summarize our findings and discuss their implications (Sec. \ref{chap4}).

\section {METHODS}
\label{chap2}
\subsection {Model Definition}
\label{chap2A}
\begin{figure}
\centering
\setlength{\abovecaptionskip}{0cm}
\includegraphics[scale=0.17]{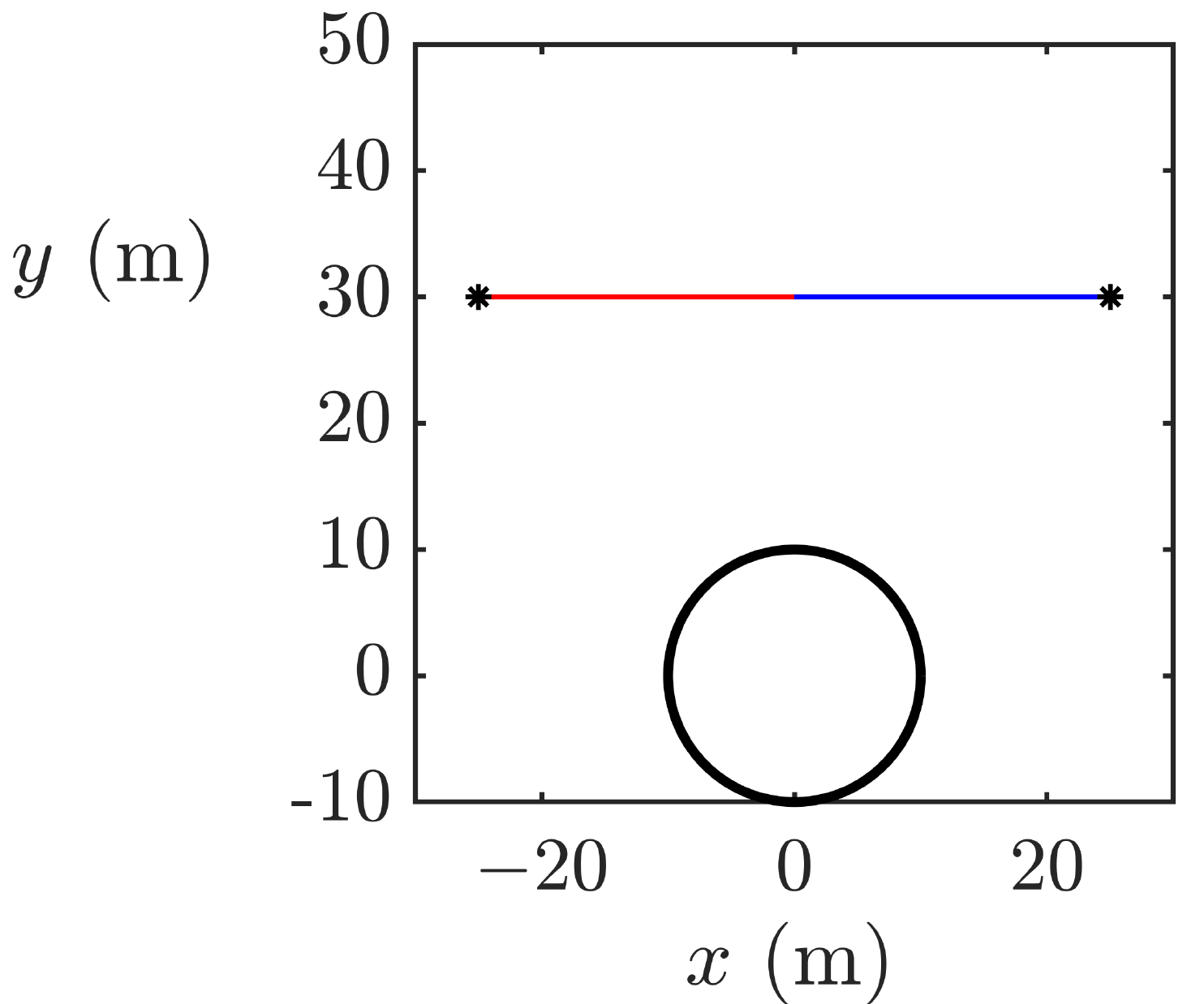} \caption{{\bf Schematic of the 2-D model layout} A screenshot of the simulation at initial state. The circle represents the body, the blue/red line represents the left/right beam, the orange arrow represents the propulsive force, and the gray arrow represents the random force that oscillates laterally.}
\label{fig:1}
\end{figure}

Our simplistic 2-D model of body-obstacle interaction (Fig.\ref{fig:1}) consists of three rigid bodies on a horizontal plane: a circular disc representing the locomotor body (mass $M$ = 1 kg, radius $R$ = 10 m), and two plates with viscoelastic revolute joints at their far ends (mass $m$ = 0.1 kg, length $L$ = 25 m, moment of inertia about joint $I$ = 20.83 kg$\cdot$m$^2$). Hereafter, we refer to them as beams. The beams align with each other when unloaded, forming a closed ``gate''. For simplicity, we assumed that the body surface is frictionless. Thus, body-beam contact results only in forces normal to the body surface, which generates no torque. Thus, body orientation remains unchanged. In addition, we assumed that 2-dimensional body-beam collisions are inelastic with partial energy loss. We used coefficient of restitution $CoR$ to measure the degree of elasticity, defined as the ratio of final to initial relative normal velocities between the body and beams during each collision. Here we tuned $CoR$ by observing different trials and chose a value of 0.8 to conserve a large portion of mechanical energy so that traversal still occurs even after a large number of collisions. Finally, for simplicity, we assumed that there is no friction against the ground (although it can be added--see Sec. \ref{chap3C}, which should only quantitatively but not qualitatively change our results).

Besides forward propulsion, animals with a sprawled leg posture like cockroaches can generate substantial lateral forces during locomotion \cite{schmitt2000mechanical}. In addition, animals can randomly change their movement direction during locomotion \cite{moore2017unpredictability}. Considering these and for simplicity, self-propulsion of the locomotor body was modeled by a constant forward propulsive force $F_{prop}$ and an oscillating random lateral force $F_{rand}$ (Fig. \ref{fig:1} red and green arrows) with a standard Gaussian distribution:\par
\begin{equation}
    F_{rand}=Rm\times rand \label{Eqn1}
\end{equation}

\noindent where $Rm$ is a constant characterizing the magnitudes of the random forces and $rand\sim \mathcal N(0,1)$, a normal distribution with a mean of 0 and a standard deviation of 1.

Because the body does not rotate, forward and lateral forces are always in the $y$- and $x$- directions, respectively. Applying Newton's second law in these two directions, the equations of motion of the body are:
\begin{equation}
\text{Body: }M \vec a=\overrightarrow{F_{prop}}+\overrightarrow{F_{rand}}-\sum \overrightarrow{F_{resis}^j} \label{Eqn2}
\end{equation}
where $M$ is body mass, $a$ is body acceleration, and $F_{resis}^j$ is the resistive force from the $j^{th}$ beam, with $j=1, 2$ for the left and right beams, respectively.

Applying Newton's second law to the beam:
\begin{equation}
\text{Beam: }I\beta_j=F_{resis}^j-k_j\theta_j-d_j\omega_j \label{Eqn3}\end{equation}
where $I$ is beam moment of inertia about its joint, $\theta_j,\omega_j,\beta_j$ are the orientation, angular velocity, and angular acceleration of the $j^{th}$ beam, and $k_j$ and $d_j$  are the torsional stiffness and damping coefficient of the $j^{th}$ beam joint.

\subsection {Interaction Dynamics}
\label{chap2B}
We used the Euler method to numerically integrate forward in time to calculate the dynamics of the system. A time step of 0.004 s was chosen to achieve good numerical accuracy while maintaining computation efficiency. The run-time of each trial varied from 5 to 20 min on a lab workstation. The key part of dynamics is to determine the interaction between the body and beams. We developed two complementary methods to model interaction dynamics for two different scenarios: a collision method and a constraint method. The collision method models repeated collisions between body and beams after initial contact, which occur due to the competing forward propulsive force and backward beam resistive forces. However, as these collisions dissipate energy, the duration of single collision decreases and eventually becomes smaller than the time step,  leading to a significant increase in numerical error \cite{ames2005sufficient}. In this case, we assumed that the interacting body and beam have no relative motion in the direction normal to the contact point but can move relatively in the tangential direction and developed the constraint method to describe interaction dynamics.

\subsubsection{Collision method}

When calculating collision occurring over an infinitesimal time, we considered the body and beam as a system with the center of rotation at the beam joint. The system is subject to a finite external beam joint torque and external forces $F_{prop}, F_{rand}$. After multiplying the finite values with an infinitesimal time, the external momentum impulse is negligible. In addition, there is a reaction force acting on the fixed beam joint, which has a zero moment arm. Therefore, the angular momentum of the system about the instantaneous center of rotation is conserved during a collision.

The calculation of collisions depends on the geometric relationship between body and beam. During traversal, there are two different contact cases: tangential contact and point contact.

\begin{figure}[H]
\centering
\setlength{\abovecaptionskip}{0cm}
\includegraphics[scale=0.18]{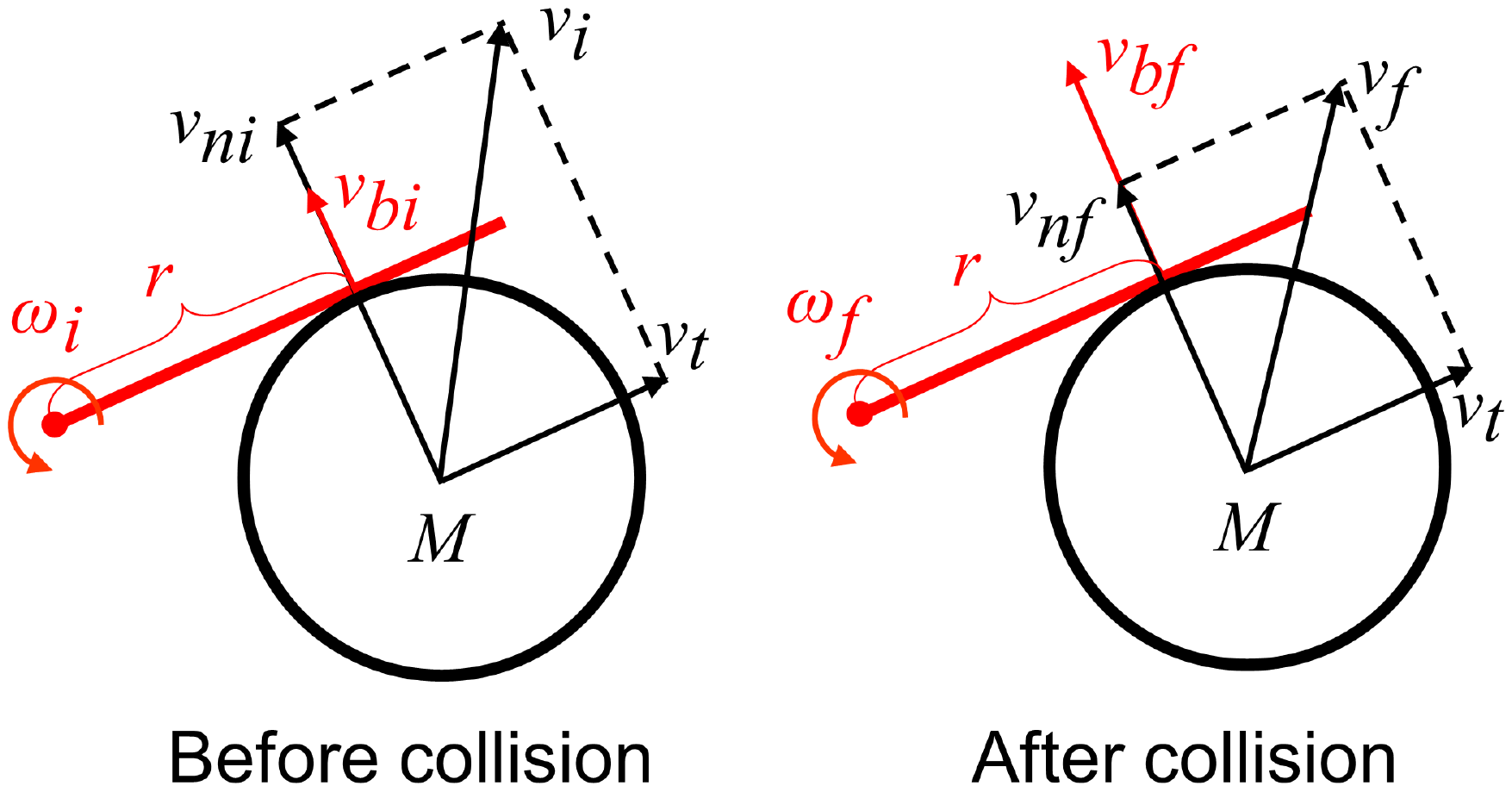} \caption{\bf Schematics of the tangential contact between the body and left beam before and after a collision.}
\label{fig:2}
\end{figure}

In the tangential contact case (Fig. \ref{fig:2}), the beam is tangential to the body surface at the point of contact. The normal direction of the body surface at the point of contact is perpendicular to the beam. Applying conservation of angular momentum to the body-beam system:
\begin{equation}
    Mrv_{ni}+I\omega_i=Mrv_{nf}+I\omega_f
    \label{Eqn4}
\end{equation}
In addition, applying the definition of CoR to the tangential contact case:
\begin{equation}
    CoR=\frac{v_{nf}-r\omega_f}{r\omega_i-v_{ni}}
    \label{Eqn5}
\end{equation}
where $r=\sqrt{x^2+y^2-R^2}$ is the distance from beam joint to contact point, $\omega_i$ and $\omega_f$ are beam angular velocities before and after collision, $v_{bi}$ and $v_{bf}$ are local velocities of the beam at the contact point before and after collision, $v_{i}$ and $v_{f}$ are body velocities before and after collision, $v_{ni}$, $v_{nf}$, $v_{ti}$, and $v_{tf}$ are their projections in the direction normal and tangential to the beam. Note that $v_{ti}$ = $v_{tf}$ because there is no interaction force in the tangential direction from our frictionless body assumption. Using Eqns. 4 and 5, we can solve the dynamics of the collision in the tangential contact case.

\begin{figure}[H]
\centering
\includegraphics[scale=0.20]{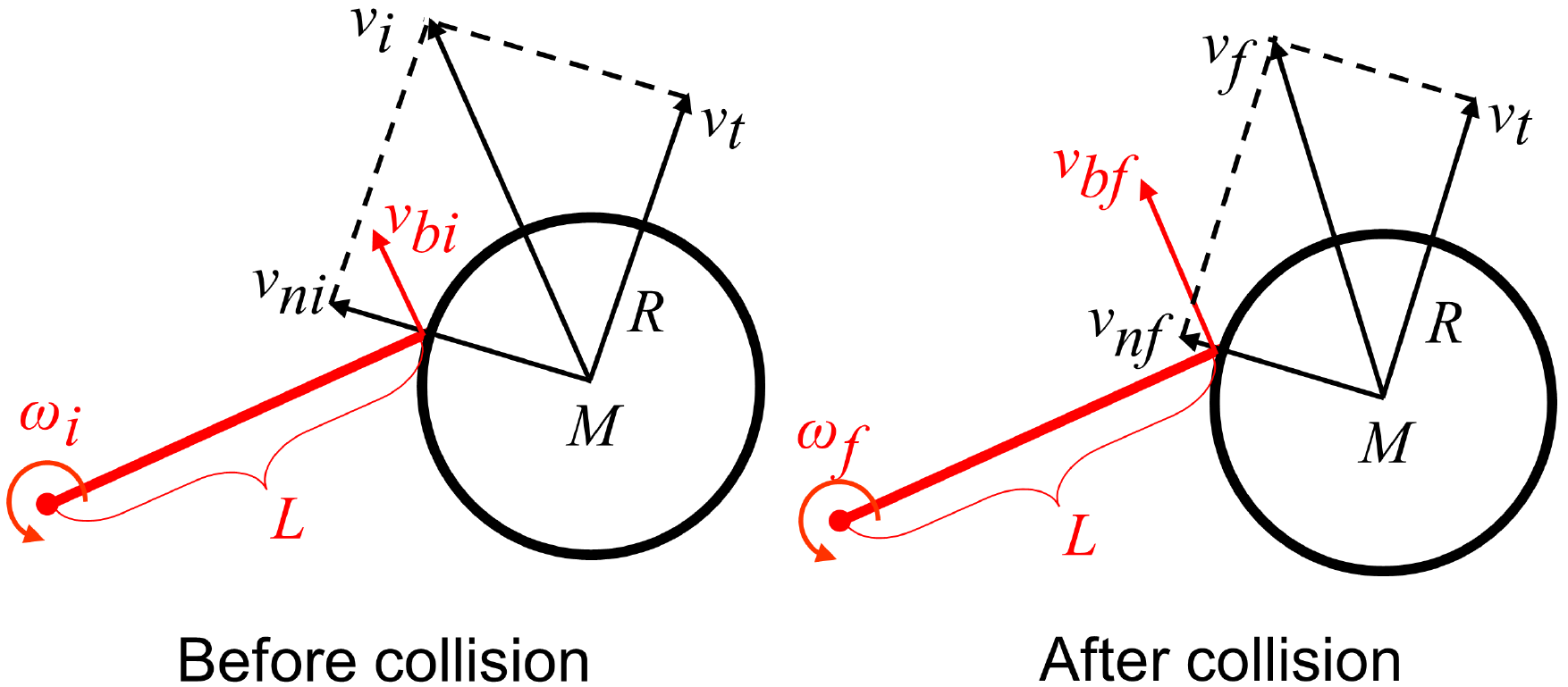} \caption{{\bf Schematics of the point contact case before and after a collision.}}
\label{fig:3}
\end{figure}

In the point contact case (Fig. \ref{fig:3}), the free end of the beam contacts the body surface. Because the normal direction of the body surface at the contact point is no longer perpendicular to the beam, it is more complex to formulate a $CoR$ expression in scalar form. Considering that its collision is small and only happens in the short duration before detachment, we assumed the collision to be perfectly elastic in this case for simplicity. Thus, mechanical energy is assumed to be conserved at the collision time step:
\begin{equation}
  \Delta E=\frac{1}{2}M\left({v_f}^2-v_i^2\right)+\frac{1}{2}I\left(\omega_f^2-\omega_i^2\right)=0
    \label{Eqn9}
\end{equation}

Also the system complies with the conservation of angular momentum before and after collision:
\begin{equation}
  \Delta P=M(v_f-v_i)=F\Delta t
    \label{Eqn6}
    \vspace{-0.5cm}
\end{equation}
\begin{equation}
  \Delta L_{beam}=I(\omega_f-\omega_i)=-r\times F\Delta t
    \label{Eqn7}
        \vspace{-0.5cm}
\end{equation}
\begin{equation}
  \Delta L_{sys}= r\times \Delta P + \Delta L_{beam} =0 
    \label{Eqn8}
\end{equation}
where $\Delta P$ is the change of body momentum, $v_i$ and $v_f$ are body velocities before and after the collision, $\Delta L_{beam}$ is the change of angular momentum of the beam before and after the collision, $F$ is the interaction force acting on the body from the beam, $r$ is the distance from the beam joint to contact point, and $\Delta L_{sys}$ is the change of angular momentum of the system, which is zero at collision. The velocities after the collision can be solved from those before using Eqns. \ref{Eqn9}-\ref{Eqn8}.

\subsubsection{Constraint method}

The constraint method is based on Newton’s second law and the geometric constraints that the body and beam have the same normal velocity at the contact point. It also differs between the two contact cases.

In the tangential contact case, the velocity constraint is:
\begin{equation}
\omega r=-\sin \theta v_x+ \cos \theta v_y
    \label{Eqn10}
\end{equation}
Taking time derivative of both sides of Eqn. \ref{Eqn10}, we have:
\begin{equation}
\begin{split}
&\beta r+sin\theta a_x-cos\theta a_y+\left(\frac{x}{r}+cos\theta\right)\omega v_x \\
&+\left(\frac{y}{r}+sin\theta\right)\omega v_y =0
    \label{Eqn11}
\end{split}
\end{equation}
where $(x, y)$ is the center of mass position measured from the beam joint, $(v_x,v_y)$ is body velocity, and $(a_x,a_y)$ is the body acceleration.

In the point contact case, the velocity constraint is: 
\begin{equation}
\omega L\left(\sin\theta x-\cos\theta y\right)=\left(L\cos\theta-x\right)v_x+\left(L\sin\theta-y\right)v_y
    \label{Eqn12}
\end{equation}
Taking time derivative of both sides of Eqn. \ref{Eqn12}, we have:
\begin{equation}
\begin{split}
&\beta L\left(\sin\theta x-\cos\theta y\right)= \\
&\left(L\cos\theta-x\right)a_x+\left(L\sin\theta-y\right)a_y
-{v_x}^2-{v_y}^2 \\
&+2L\omega\left(\cos\theta v_y-\sin\theta v_x\right)-L\left(\cos\theta x+\sin\theta y\right)\omega^2
\label{Eqn13}
\end{split}
\end{equation}
where $L$ is beam length.

\begin{figure}
\centering
\includegraphics[width=0.4\textwidth]{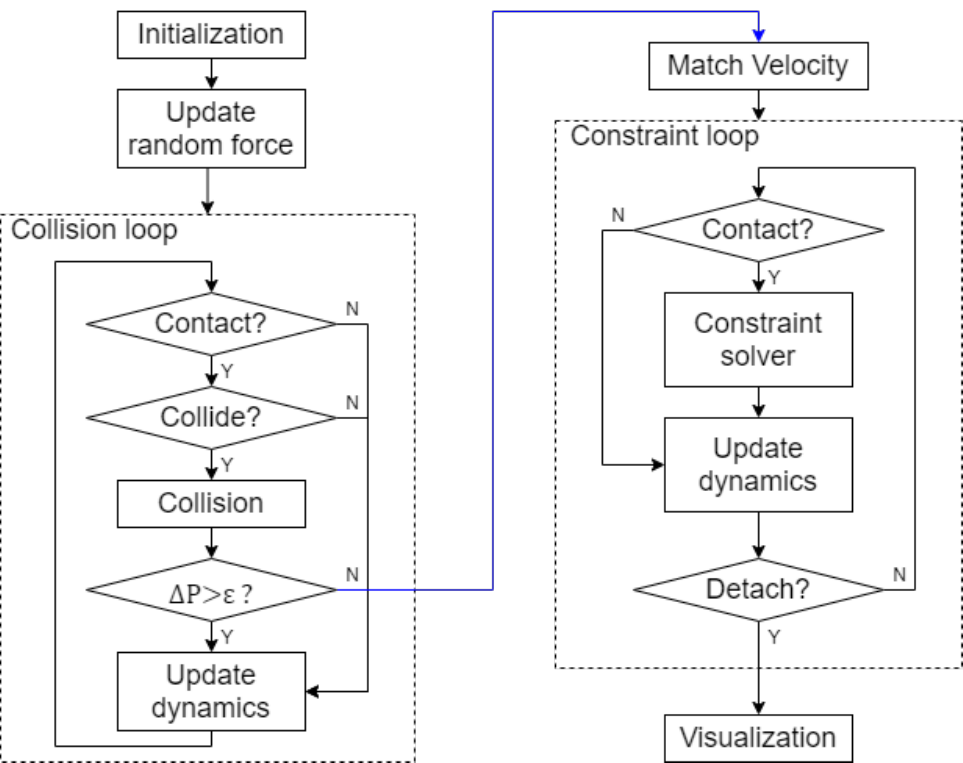} \caption[{\bf A flowchart of the simulation program. }]{{\bf A flowchart of the simulation program.}}
\label{fig:4}
\end{figure}

In the simulation, the collision and constraint methods are integrated following the workflow shown in Fig. \ref{fig:4}. The simulation first runs the collision method in loops. As the change of the body's momentum during each collision $\Delta P$ decreases below a small threshold $\varepsilon=0.04$ kg$\cdot$m/s, i.e., collisions become small enough and exchange little momentum, the simulation switches to the constraint method in loops. During the switch, beam angular velocities are updated to satisfy the initial velocity constraint of the constraint method.

\subsection{Potential Energy Landscape}
\label{chap2C}

The potential energy landscape is a modeling approach to model the conservative forces during locomotor-terrain interaction over relevant degrees of freedom. Without knowledge of non-conservative forces that are often difficult to measure, it provides a useful approach for understanding how the system may or may not move on the landscape, as long as potential energy landscape plays a major role in shaping dynamics \cite{othayoth2021locomotor}. In our system, potential energy arises from elastic torsional joints of the beams:
\begin{equation}
E_{beam}=\frac{1}{2}k_L\theta_1^2+\frac{1}{2}k_R\theta_2^2
    \label{Eqn14}
\end{equation}
where $E_{beam}$ is the total beam potential energy, $k_L/k_R$ is the elastic stiffness of the L/R beam, and $\theta_i$ is the deflection angle of the $i^{th}$ beam, with $i=1,  2$ for the left and right beams, respectively.

\begin{figure}
\centering
\setlength{\abovecaptionskip}{0cm}
\includegraphics[scale=0.32]{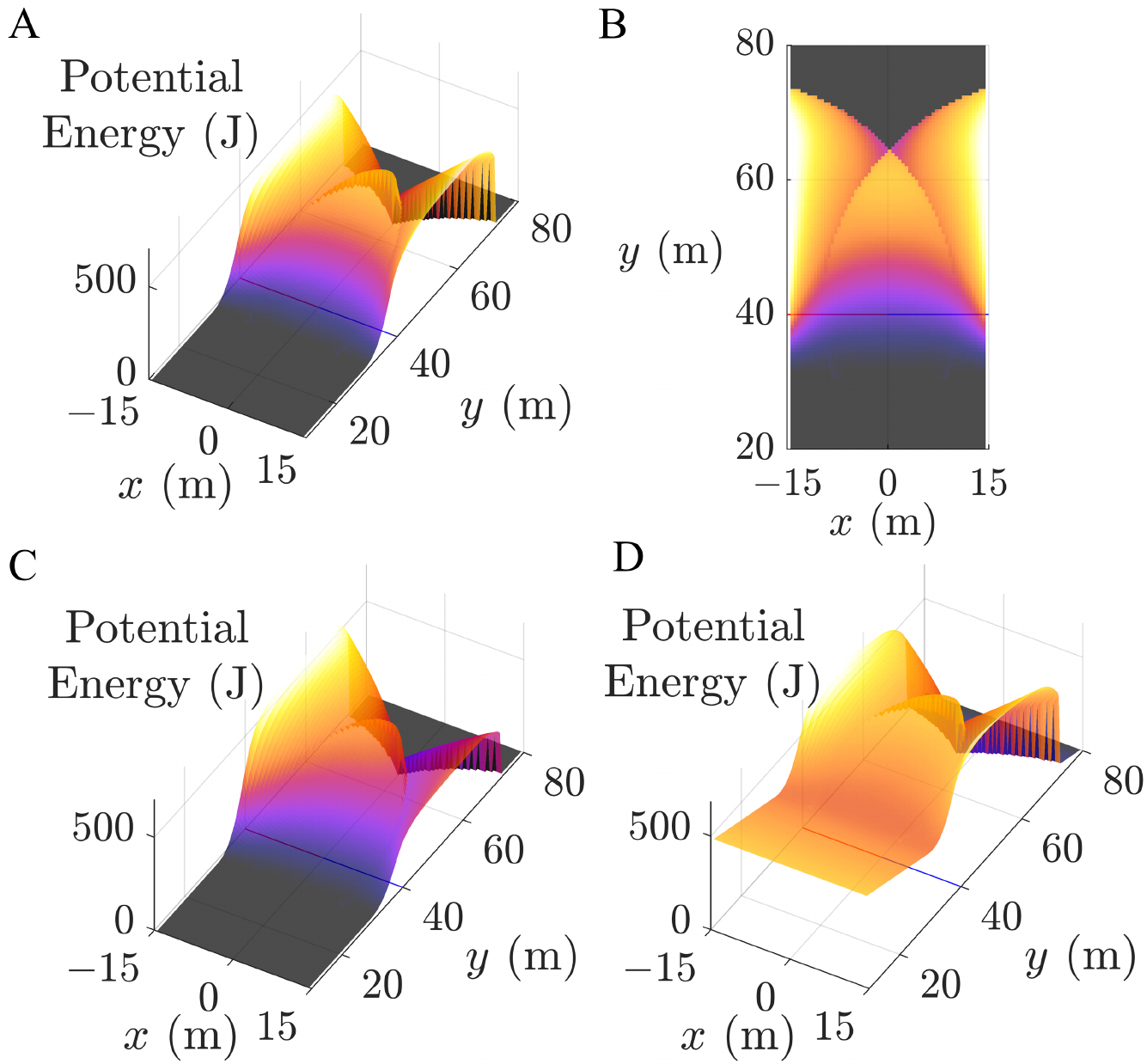} \caption{{\bf Potential energy landscape for the beam obstacle traversal.} \textbf{(A)} Landscape with only the elastic energy. $k_L=k_R$ = 400 N$\cdot$m/rad. \textbf{(B)} A top view of the landscape in A. \textbf{(C)} An asymmetric landscape with $k_L$ = 500 N$\cdot$m/rad and $k_R$ = 250 N$\cdot$m/rad. \textbf{(D)} Landscape with sum of elastic energy and propulsive conservative energy. $k_L=k_R$ = 400 N$\cdot$m/rad.}
\label{fig:5}
\vspace{0.3cm}
\end{figure}

The landscape is calculated as a 3-D surface $PE=f(x,y)$ over a mesh grid of $\delta x=\delta y=0.5$ m. At each grid point, we assumed that beams are always deflected forward and contact the body if the body is within the range of beam radius. During traversal, the body pushes forward and deflects the beams, resulting in two potential energy barriers on each side (Fig. \ref{fig:5}AB). The two barriers overlap in the middle area, where the body can interact with both beams. Fig. \ref{fig:5}C is an example of asymmetric landscape with different $k_L$ and $k_R$.

Because we assumed a constant forward $F_{prop}$, the potential energy landscape can also include conservative potential energy from $F_{prop}$.  The summed potential energy is:
\begin{equation}
E_{prop}+E_{beam}=F_{prop}\left(y_0-y\right)+\frac{1}{2}k_L\theta_1^2+\frac{1}{2}k_R\theta_2^2
    \label{Eqn:15}
\end{equation}
where $E_{prop}$ is the potential energy from $F_{prop}$. The zero level of $E_{prop}$ was defined at the anterior boundary $y_0$ = 60 m.

A basin that spans across the $x$ direction in front of the beams (around $y$ = 20 m) emerges after including the energy of $F_{prop}$ of a modest magnitude in the potential energy landscape (Fig. \ref{fig:5}D). This helps us understand how the landscape is tilted by the propulsive force. Within insufficient $F_{prop}$, the locomotor body would be trapped in this basin and not traverse. Adding kinetic energy fluctuation induced by the random force can help the body escape the horizontal basin. With sufficient $F_{prop}$, the landscape is so heavily tilted that the basin disappears, and the locomotor should always traverse.

\begin{figure}
\centering
\setlength{\abovecaptionskip}{-0cm}
\setlength{\belowcaptionskip}{0cm}
\includegraphics[width=0.48\textwidth]{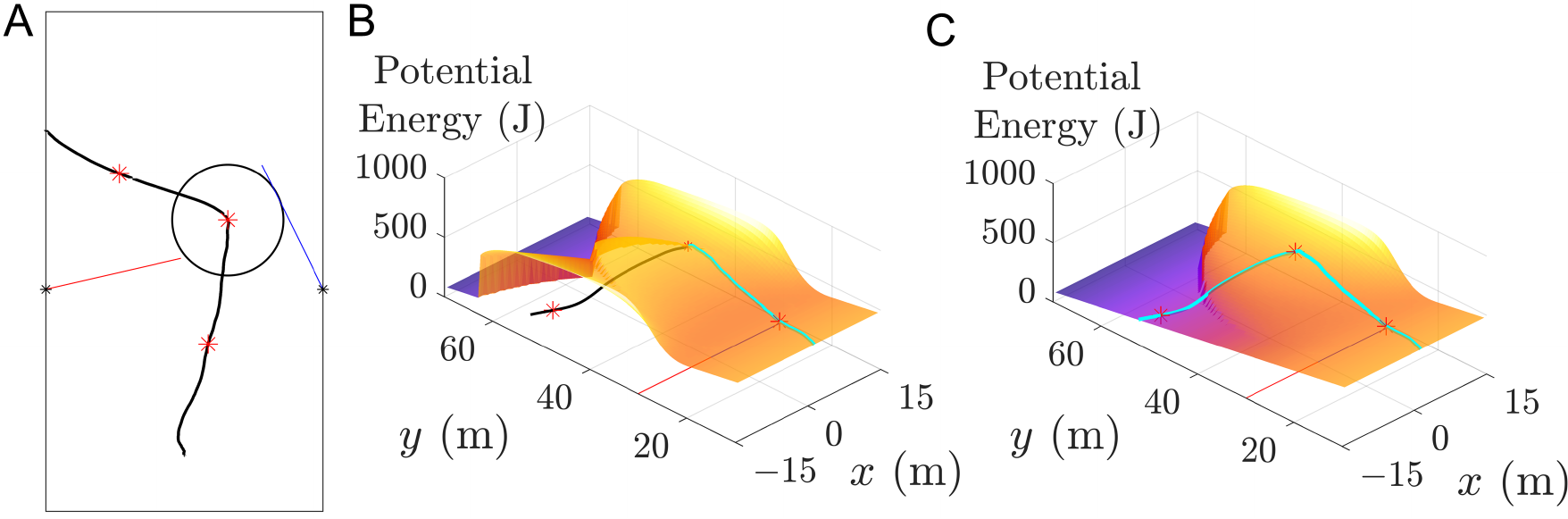} \caption{{\bf An example of trajectory penetrating the landscape surface.} Points with $\ast$ show the mapping between 2-D position and landscape. \textbf{(A)} The screenshot in simulation. \textbf{(B)} The trajectory plotted over the normal landscape. \textbf{(C)} The trajectory plotted over the landscape without the inactive barrier.}
\label{fig:6}
\end{figure}

The evolution of the potential energy of the system in the simulation traversal trials can be plotted over the landscape to visualize how the body-beam interaction influences body motion. Typically, the trajectory of the system potential energy is above the landscape surface due to the beam inertia. However, sometimes the trajectory penetrates the landscape surface. An example is shown in Fig. \ref{fig:6}B. In this case, the left beam lost contact with the body and returned its initial orientation (Fig. \ref{fig:6}A). Because the body is still within the radius of the beam, our landscape calculation that always assumes beams being deflected forward over-estimates the potential energy. If we only consider the landscape from the right beam that is deflected, this artifact is removed (Fig. \ref{fig:6}C). See \href{https://youtu.be/jR6bv94KuuE}{multimedia material} part 3 for an example video.

\subsection{Multi-Obstacle Field Interaction}
\label{chap2D}

\begin{figure}
\centering
\includegraphics[scale=0.20]{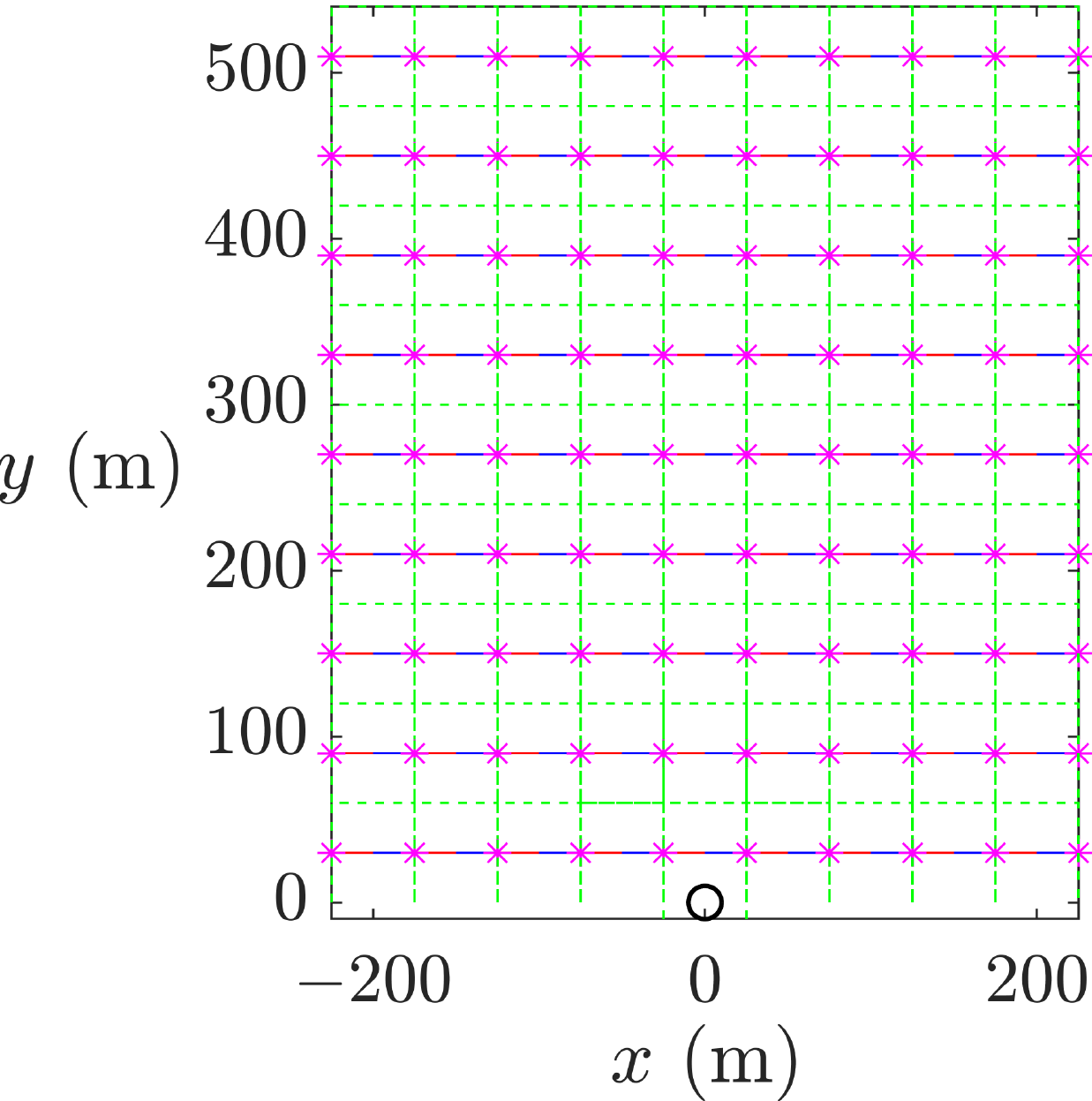} \caption{{\bf A 9 $\times$ 9 lattice obstacle field.} Green dashed line show boundary of the region of each pairs of beams. $\ast$ show beam joints. Blue and red line segments show left and right beams of each pair.}
\label{fig:7}
\end{figure}

Using the body-beam interaction model above with a single pair of beams (a gate), we can create a large cluttered obstacle field by composing multiple pairs of beams (gates) in a lattice arrangement. Fig. \ref{fig:7} shows a 9 $\times$ 9 gate obstacle field used for further simulation. Except for the first gate region which uses a set of initial system states ($[x,y,v_x,v_y]$), the system state input of each gate region comes from the output of the previous gate region. An index $i=[i_x,i_y]$ was used to track which gate region is being activated. For this 9 $\times$ 9 gait obstacle field, $i_x \in [-4,4]$ and $i_y \in [0,8]$. Finally, the main program outputs a trajectory restricted in a single gate region, which is a series of local positions $P_{local}=[x, y]$, where $x\in [-25,25]$ m, $y\in[0,60]$ m. Later in the visualization process, according to the gate index $i$, $P_{local}$ are converted to the global position data $P_{global}=[X,Y]$, where $X\in [-225,225]$ m, $Y\in[0,540]$ m. See \href{https://youtu.be/jR6bv94KuuE}{multimedia material} part 4 for a simulation example.


The pseudo code:
\begin{algorithm}
        \renewcommand{\thealgorithm}{}
        \caption{Gate lattice simulation}
        \begin{algorithmic}[0] 
            \State Initialization, $i \gets [0,0], t \gets 0$
            \While{$i_x^{min} \le i_x \le i_x^{max}$ \textbf{and} $i_y^{min} \le i_y \le i_y^{max}$}
                
                \State Single gate simulation, $t=t+\Delta t$
                \If {Reached left boundary, $x < -25$ m}
                    \State $i_x \gets i_x -1$, $x \gets x +25$ 
                \ElsIf{Reached right boundary, $x > 25$ m}
                    \State $i_x \gets i_x +1$, $x \gets x -25$ 
                \ElsIf{Reached top boundary, $y > 60$ m}
                    \State $i_y \gets i_y+1$, $y \gets y -60$ 
                \EndIf
                \State $P_{local}(t)=[x,y],i(t)=[i_x,i_y]$
            \EndWhile
            \State\Return{$\left\{P_{local}\right\},\left\{i\right\}$}
        \newline
        \Function{mat2avi}{$\left\{P_{local}\right\},\left\{i\right\}$}
            \State Convert local position data, $P_{global} \gets P_{local}$
            \State\Return {video.avi}
        \EndFunction
        \end{algorithmic}
    \end{algorithm}
    
All the modeling calculations and simulations were performed in MATLAB R2019b.
    
\section {RESULTS}
\label{chap3}
\subsection{Traversing symmetric beams}
\label{chap3A}
We first studied the probability of traversing a pair of symmetric beam obstacles under self-propulsive and random forces. The body starts at an initial forward velocity $v_0$, accelerates for a short distance $d_{acc}$ before making initial contact with beams. The criterion of traversing successfully is that the body can reach $y$ = 65 m, where the body is out of the range of beams (\href{https://youtu.be/jR6bv94KuuE}{multimedia material}, part 1). Failure resulted from: (1) the body gets stuck by the beams (\href{https://youtu.be/jR6bv94KuuE}{multimedia material}, part 2), (2) the body deviates from the desired track and exits from the side boundaries ($x \le$ -25 m or $x \ge$ 25 m), and (3) the body did not reach $y$ = 65 m within maximal iterations.

From the potential energy landscape (Fig. \ref{fig:5}A), the locomotor body needs to have sufficient kinetic energy to overcome a potential energy barrier to traverse \cite{othayoth2020energy}. Because higher propulsive force can tilt the landscape, facilitating to overcome barriers by lowering them, while the random force adds more kinetic energy fluctuation in the system. Thus, we hypothesized that the probability of traversing a pair of symmetric beam obstacles increases with propulsive force and random force.

\begin{figure}
\centering
\includegraphics[width=0.40\textwidth]{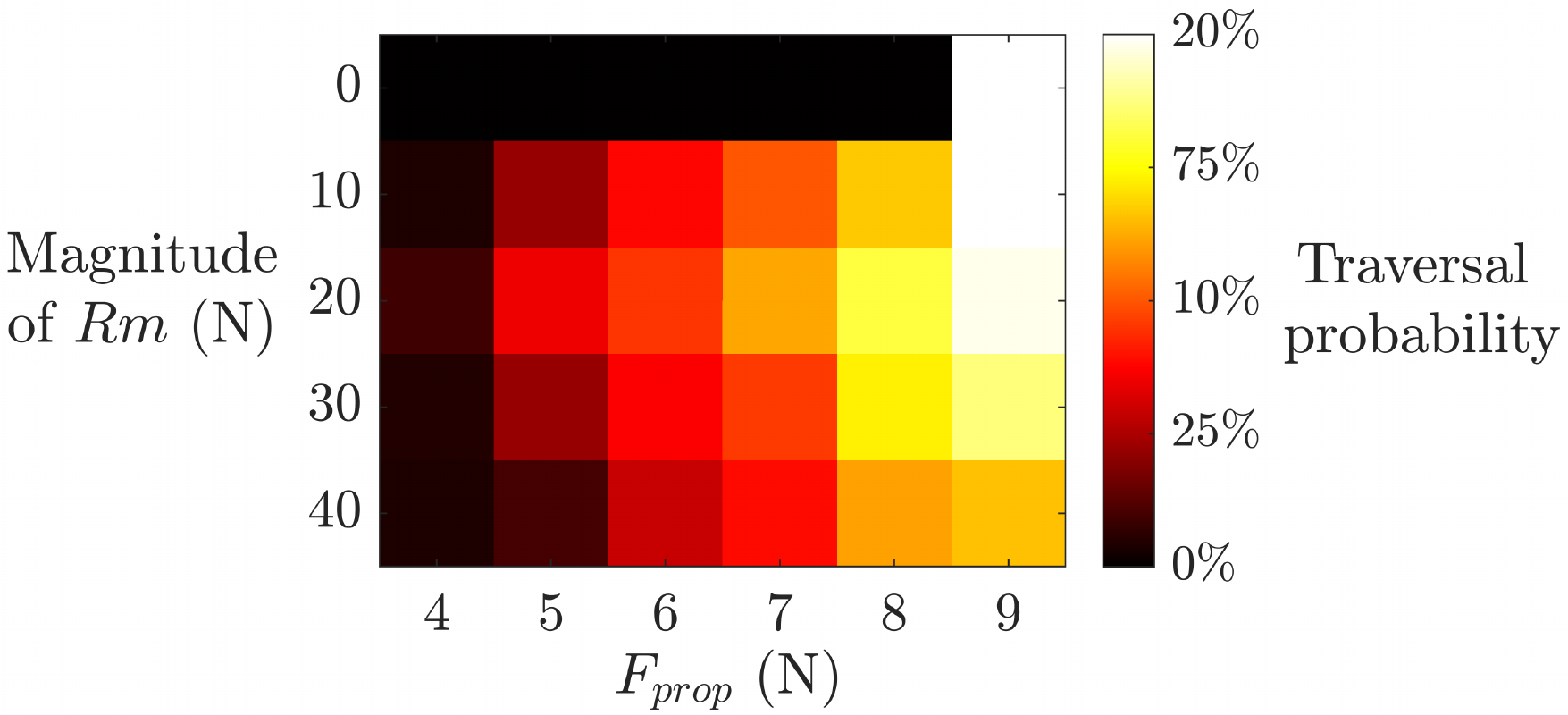} \caption{{\bf Probability of traversing a pair of symmetric beams as a function of propulsive force and random force magnitude.}}
\label{fig:8}
\end{figure}

To test this hypothesis, we varied propulsive force $F_{prop}$ and the magnitude of random force $Rm$ using system parameters in Table \ref{tab2} and ran 100 trials for each given ($F_{prop}, Rm$) to obtain traversal probability (Fig. \ref{fig:8}). For a given $Rm$, traversal probability increased with propulsive force. However, for a given $F_{prop}$, traversal probability did not always increase with random force. A random force of $Rm$ up to 20 N increased traversal probability. However, as $Rm$ increased beyond 20 N, traversal probability decreased. Examination of simulation videos revealed that this was because more trials exited the side boundaries of the gate region before arriving at the beams. Thus, a sufficient but not excessive lateral random force helps traverse symmetric beam obstacles.

\begin{table}
\setlength{\abovecaptionskip}{-0.1cm}
\setlength{\belowcaptionskip}{0cm}
\caption{Symmetric test configuration\strut}
\label{tab2}
\begin{center}
\begin{tabular}{cc}
\hline
Parameter & Value \\
\hline
Propulsive force $F_{prop}$ & 4,5,6,7,8,9 N \\
Random force magnitude $Rm$ & 0,10,20,30,40 N \\
Beam stiffness $k_L,k_R$ & 400 N$\cdot$m/rad \\
Beam damping $d_1,d_2$ & 50 N$\cdot$s/rad \\
Initial forward velocity $v_0$ & 0 m/s \\
Coefficient of Restitution $CoR$ & 0.8 \\
Oscillator frequency $f$ & 50 Hz \\
Acceleration distance $d_{acc}$ & 20 m \\
Max iterations  & 3000 \\ \hline
\end{tabular}
\end{center}
\end{table}

\subsection{Traversing asymmetric beams}
\label{chap3B}
Next, we studied the probability of traversing a pair of asymmetric beams with different stiffness. The potential energy landscape becomes asymmetric in this case, with a barrier lower on one side where the beam has a smaller stiffness (Fig. \ref{fig:5}C). We hypothesized that the probability of traversing on the side of lower stiffness increases with the level of asymmetry.  

\begin{figure}[H]
\vspace{0.2cm}
\centering
\setlength{\abovecaptionskip}{-0cm}
\setlength{\belowcaptionskip}{0cm}
\includegraphics[width=0.48\textwidth]{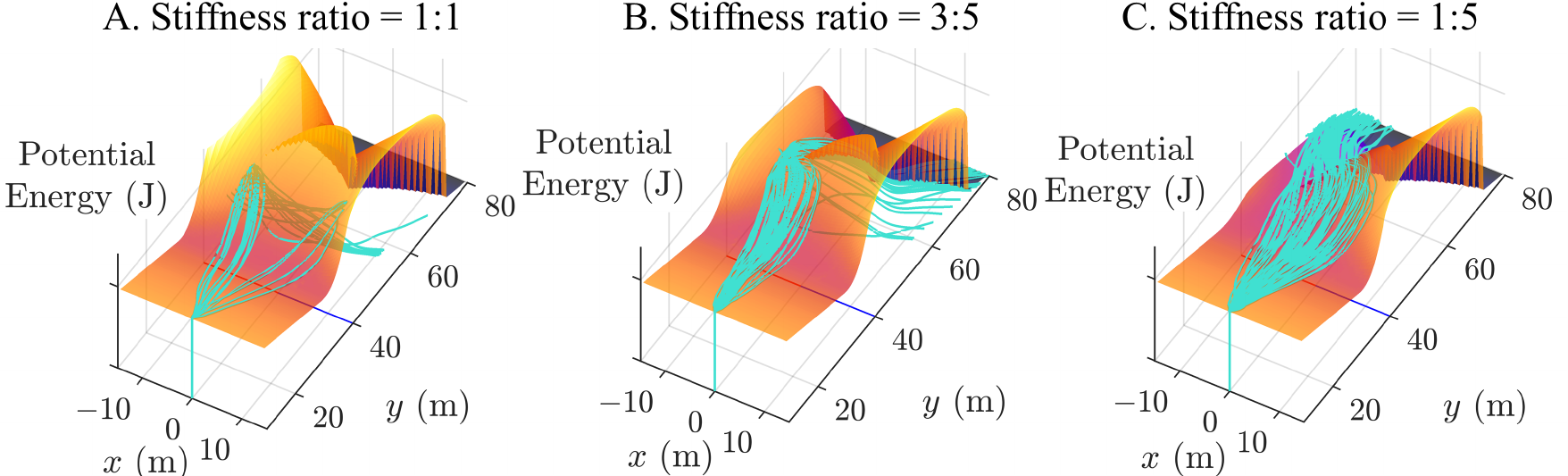} \caption{{\bf Successful trajectories on the landscape in asymmetric test.} Self-propulsive force = 7 N, random force magnitude = 10 N, $k_R=500$ N$\cdot$m/rad. \textbf{(A)} $k_L=500$ N$\cdot$m/rad. \textbf{(B)} $k_L=300$ N$\cdot$m/rad. \textbf{(C)} $k_L=100$ N$\cdot$m/rad.}
\label{fig:9}
\end{figure}

To test this hypothesis, we varied the left beam stiffness while keeping the right beam stiffness constant and varied $F_{prop}$ and $Rm$ systematically (Table \ref{tab3}). Unlisted parameters are the same as those in Table \ref{tab2}. 
Here we selected three groups with varying stiffness ratios as examples to illustrate the typical results (Fig. \ref{fig:9}). Only successful traversal trajectories are plotted over the landscape. In the symmetric control group with both beam stiffness being 500 N$\cdot$m/rad (Fig. \ref{fig:9}A), traversal was rare with similar probability on both sides (L:13\%, R:8\%). This is because the propulsive force was barely sufficient to overcome the potential energy barriers. When the left beam stiffness was reduced from 500 to 300 N$\cdot$m/rad, traversal was frequent (52\%), with the body pushing across the left beam (Fig. \ref{fig:9}B). As the left beam stiffness further reduced to 100 N$\cdot$m/rad, traversal was almost guaranteed by pushing across the left beam (98\%, Fig. \ref{fig:9}C). 

\begin{table}
\setlength{\abovecaptionskip}{-0.1cm}
\setlength{\belowcaptionskip}{0cm}
\caption{Asymmetric test configuration}
\label{tab3}
\begin{center}
\begin{tabular}{cc}
\hline
Parameter & Value \\
\hline
Propulsive force $F_{prop}$ & 7,8N \\
Random force magnitude $Rm$ & 10,20,30,40,50,60 N \\
Left beam stiffness $k_L$ & 100:50:500 N$\cdot$m/rad \\
Right beam stiffness $k_R$ & 500 N$\cdot$m/rad \\
\hline
\end{tabular}
\end{center}
\end{table}

\begin{figure}
\centering
\includegraphics[width=0.45\textwidth]{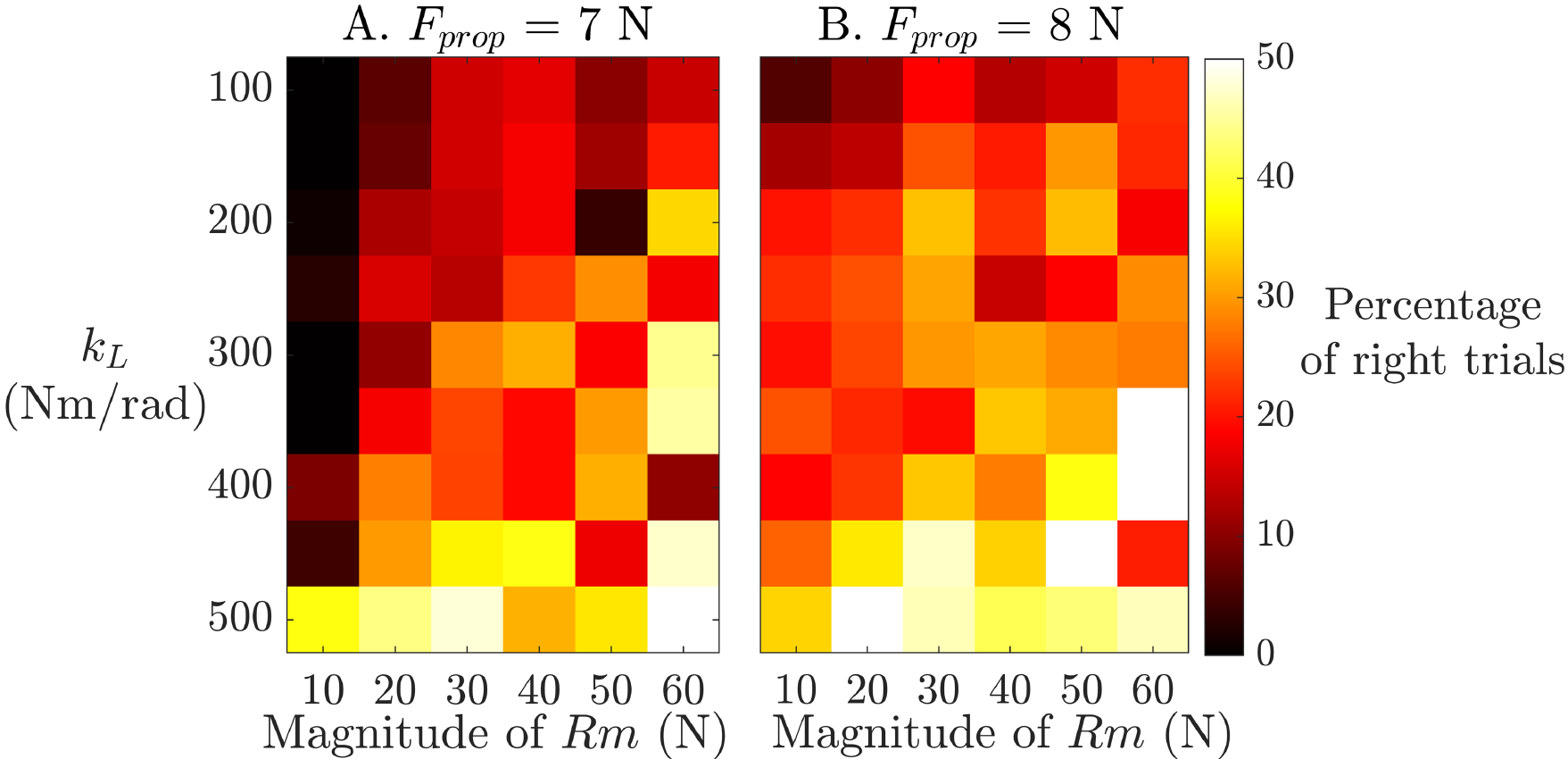} \caption{{\bf Ratio of right trials to all successful trials as a function of left beam stiffness and random force magnitude.} A darker block in the matrix indicates that there are fewer right trials than the left ones, and traversal becomes more asymmetric towards the left. A brighter block indicates the number of left and right trials are more equal.}
\label{fig:12}
\vspace{-0.05cm}
\end{figure}

For all the asymmetric tests varying $F_{prop}$ and $Rm$, we classified the trajectories into ``left'' or ``right'' types, reflecting traversal tending to which barrier side, by comparing the maximal deflected angle of the left or right beam. Then, we can quantitatively evaluate the asymmetry of trajectories by calculating the ratio of the number of trials for each type to the total number of successful traversal trials (Fig. \ref{fig:12}). For $Rm < 40$ N, as the left beam stiffness decreases, traversal becomes less frequent on the right side. However, for $Rm \ge 40$ N, this trend is still existed, but less monotonically due to the high stochasticity.  In addition, traversal probability increased with $F_{prop}$ (Fig. \ref{fig:12}A vs. B). These results well supported our hypothesis and were consistent with observations in the traversal of symmetric beams.

\subsection{Multi-Obstacle Traversal}
\label{chap3C}

\begin{figure}
\centering
\includegraphics[width=0.18\textwidth]{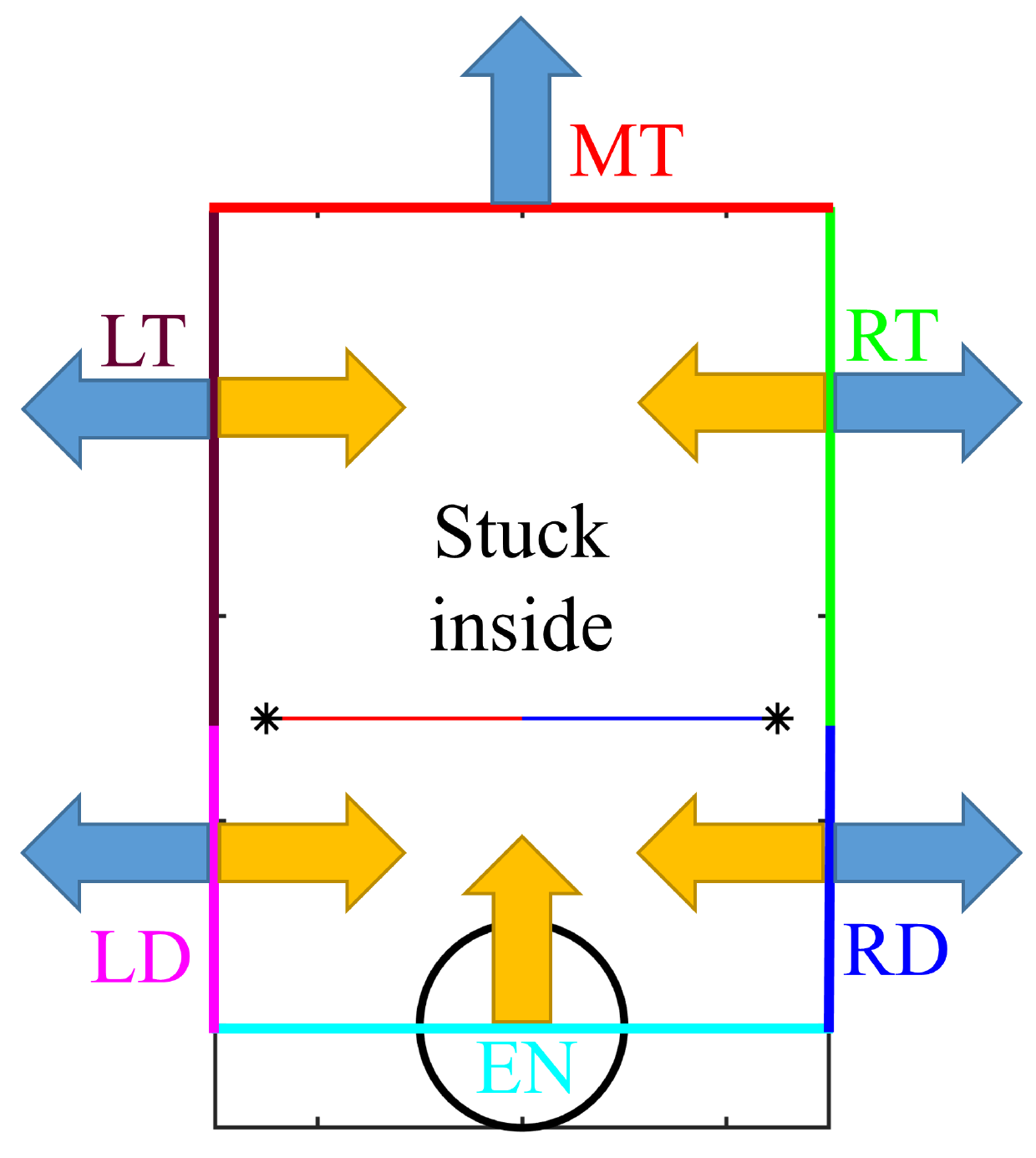} \caption[{\bf Input and output in a gate region.}]{{\bf Input and output in a gate region.} The boundary is divided into 6 segments, which are in different color. Blue arrows represent possible outputs. Orange arrows represent possible inputs.}
\label{fig:10}
\end{figure}

On the larger terrain, trajectories become more complicated and lack of patterns. Here we focused on how the locomotor body transits between adjacent gate regions. As a first step, we assumed that all the gates are symmetric with the same stiffness (400 N$\cdot$m/s). Each gate is considered as a single system with an input and an output when crossing region boundaries. Here we discretized the boundary of a gate region into 6 segments: middle traverse (MT), left traverse (LT), right traverse (RT), left deflect (LD), right deflect (RD) and enter (EN) (Fig. \ref{fig:10}). For the traversal problem, EN is only for input and MT is only for output, whereas the other four can be either input or output. Another possible output is that the body is trapped inside a gate region in between the two beams. Motion in each gate can be treated individually as a stochastic process independent of the history before entering. And we assume for fixed model parameters, there is a constant probability correspondence between input and output states. Thus, the Markov chain can be applied to describe the stochastic transition between gate regions. 

For a discrete-time Markov process, the continuous states needed to be discretized into a finite number of input and output states. Here, the state of the body when crossing the boundary is $q=[B,d,v_x,v_y]$, where $B \in $ \{EN, RD, LD, RT, LT, MT\} is the boundary index, $d$ is the position on the corresponding the boundary. $d$ is measured as the distance to several benchmark points (EN: $[0,0]$ m, LT/LD: $[-25,30]$ m, RT/RD: $[25,30]$ m, MT: $[0,60]$ m). $[v_x, v_y]$ is the velocity when crossing boundaries. To set up the state space for the Markov chain, we discretized the input states for boundaries EN, RD and LD (Table \ref{tab4}). Inputs from RT or LT rarely occur (fewer than 4\% of all trials) and they are not involved with the body-beam interaction. Thus, for simplicity, we did not further discrete RT and LT like the other boundaries to improve computational efficiency. In total, there are 87 input states, $[I_1,\ldots,I_{87}]$, and 88 output states, $[O_1,\ldots,O_{88}]$, with $O_{88}$ being the state of being trapped in a gate.

\begin{scriptsize}

\begin{table}[t]
\setlength{\abovecaptionskip}{-0cm}
\setlength{\belowcaptionskip}{0cm}
\caption{Discretization of Input States}
\label{tab4}
\begin{tabular}{|c|c|c|c|}
\hline
       & EN   & RD & LD \\ \hline
$d$ (m)      & \begin{tabular}[c]{@{}c@{}}{[}-25,-15{]}, {[}-15,-5{]},\\ {[}-5,5{]}, {[}5,15{]}, \\ {[}15,25{]}\end{tabular} & {[}10,20{]},{[}20,30{]} & {[}10,20{]},{[}20,30{]}\\ \hline
$v_x$ (m/s)    & \begin{tabular}[c]{@{}c@{}}{[}-25,-15{]}, {[}-15,-5{]},\\ {[}-5,5{]}, {[}5,15{]}, \\ {[}15,25{]}\end{tabular} & \begin{tabular}[c]{@{}c@{}}{[}-30,-20{]},{[}-20,-10{]},\\ {[}-10,0{]}\end{tabular}                         & \begin{tabular}[c]{@{}c@{}}{[}0,10{]},{[}10,20{]},\\ {[}20,30{]}\end{tabular}  \\ \hline
$v_y$ (m/s)    & {[}10,20{]} & \begin{tabular}[c]{@{}c@{}}{[}-20,-10{]},{[}-10,0{]},\\ {[}0,10{]},{[}10,20{]},\\ {[}20,30{]}\end{tabular} & \begin{tabular}[c]{@{}c@{}}{[}-20,-10{]},{[}-10,0{]},\\ {[}0,10{]},{[}10,20{]},\\ {[}20,30{]}\end{tabular} \\ \hline
Groups & 25& 30   & 30   \\ \hline
\end{tabular}
\vspace{0.2cm}  
\end{table}

\end{scriptsize}

For Markov Chain analysis, we need to obtain the transition matrix $M$, where $M_{i,j}$ is the probability from input $I_i$ to the output $O_j$ during a gate traversal (Eqn. \ref{Eqn16}). For our discretization, $M$ is an $87\times 88$ matrix. The matrix is obtained by running 100 trials per group with random force in the single gate simulation.
\begin{equation}
\begin{split}
\label{Eqn16}
&P_{after}=[P_{o1},...,P_{o88}] \\
&=[P_{i1},...,P_{i87}] 
\begin{bmatrix}
M_{1,1} & \cdots & M_{1,88} \\
\vdots & \ddots & \vdots \\
M_{87,1} & \cdots & M_{87,88} \\
\end{bmatrix}
=P_{before}\cdot M
\end{split}
\end{equation}

\noindent
where $P_{after}$ is the probability of output state after the traversal, $P_{before}$ is probability of input state before the traversal.

With the transition matrix, the Markov chain Monte Carlo (MCMC) method \cite{gilks1995markov} can be used to predict the consecutive transitions between gate units. Monte Carlo is a statistical simulation method relying on repeated random sampling from a probability distribution. MCMC is the combination of Markov chain and Monte Carlo. In each gate, input is known, either from the initialization or from the former gate. The probability of outputs after traversing this gate is given by multiplying the Markov chain transition matrix. We randomly pick a certain output state based on the probability distribution $P_{after}$ and use it as the input of the next gate. By repeating the Markov chain and Monte Carlo in turn, we can obtain a path across the terrain. The simulation ends after a maximal number of iterations. The pseudo code:

\begin{algorithm}
        \renewcommand{\thealgorithm}{}
        \caption{Markov Chain Monte Carlo}
        \begin{algorithmic}[0]
            \State Initialization, $P_{before} \gets [0,..,1,..,0]$
            \While{Body within terrain}
                \State Markov Chain $P_{after} \gets P_{before} M$
                \State Randomize one output $O_n$ $\gets$ Sample from $P_{after}$
                \If{$O_n$=$O_{88}$}
                Break
                \EndIf
                \State Transition to the next gate, new input $I_n \gets O_n$
                \State Reset $P_{before}$ $\gets$ zeros(1, 87)
                \State $P_{before}(n)=1$
            \EndWhile
        \end{algorithmic}
     
\end{algorithm}

Unlike our dynamic simulation, the Markov chain Monte Carlo algorithm cannot precisely predict the trajectory across the terrain. But we can compare the prediction for the finial lattices where the trials finally stop. By repeating MCMC trials, we can obtain a predicted distribution of the final lattice. To compare two methods, we have the same parameters of gates in dynamic simulation as the ones in MCMC.  Ideally, the MCMC simulation can predict the path on an infinitely large terrain. However, if the body keeps accelerating after traversing a series of gates, the locomotor velocity will keep increasing towards infinity, and it would be impractical to discretize velocity for the Markov chain analysis. To address this issue, we increased the beam damping ($d_1=d_2=50$ N$\cdot$s/rad) and added a viscous force ($\vec f = -D \vec v,  D = 0.06$ N$\cdot$s/m) between body and the horizontal plane so that the velocity can remain in the tested range (Table. \ref{tab4}). The body started at the origin of the entire terrain ($X$ = 0 m, $Y$ = 0 m) with $v_x$ = 0 m/s and $v_y$ = 15 m/s. For our $9 \times 9$ gate lattice, each MCMC trial is allowed to have a maximum of 13 steps to reach the furthest gate. But if the body exits from the terrain or becomes trapped inside a gate (i.e., reaching $O_{88}$), the trial will stop at that gate.

\begin{figure}
\centering
\includegraphics[width=0.48\textwidth]{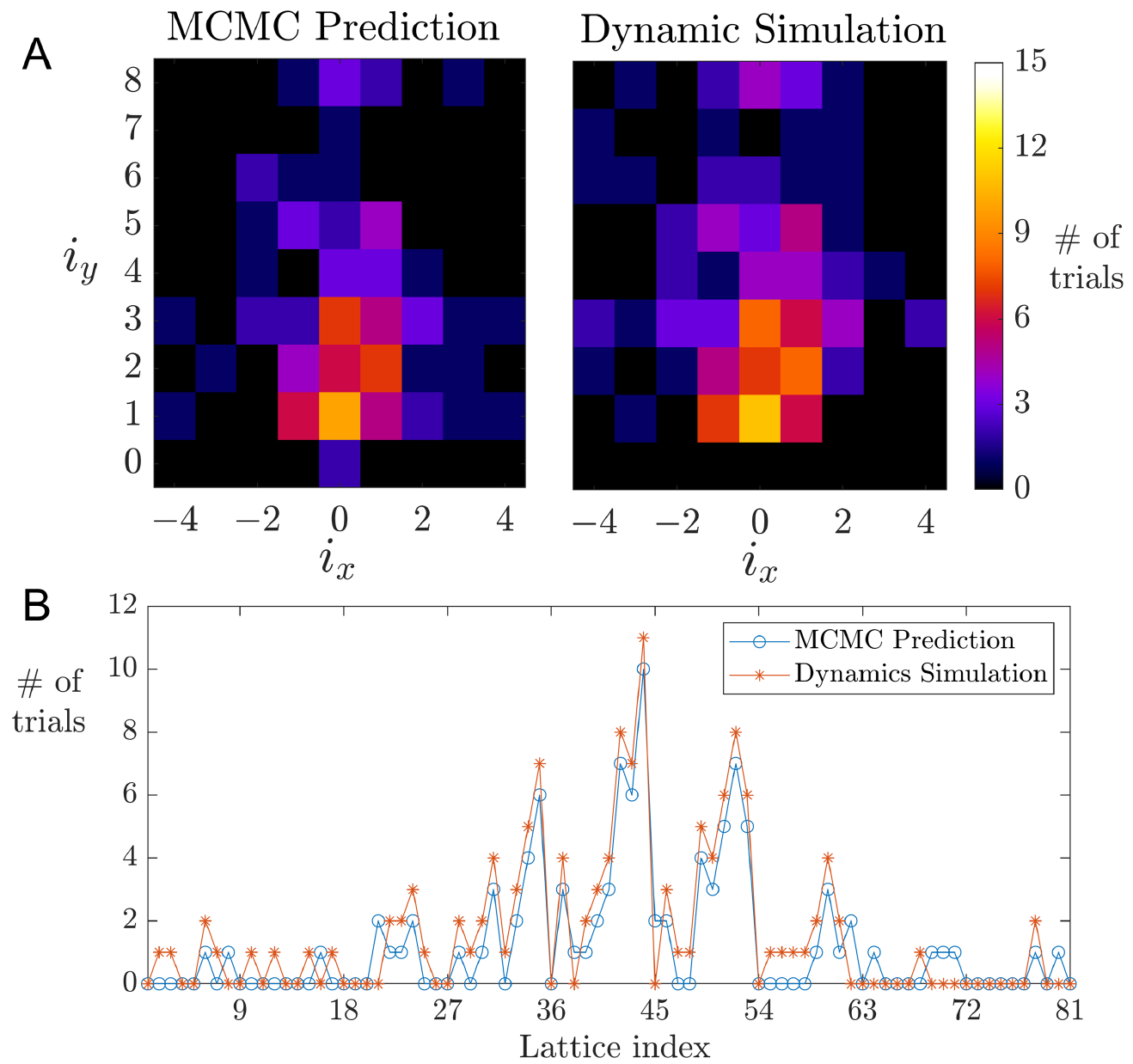} \caption{{\bf Distribution of the final location of the body within the obstacle field.} \textbf{(A)} The predicted distribution given by MCMC simulation (left) and dynamic simulation (right). \textbf{(B)} Data in A, rearranged into a more direct pair-wise comparison.}
\label{fig11}
\vspace{0.1cm}
\end{figure}

To test how well the MCMC simulation predicted traversal outcome, we ran 100 trials using both dynamic and MCMC simulation and compared the distribution of the final location of the body within the obstacle field.
The patterns are strikingly similar (Fig. \ref{fig11}). 
The correlation coefficient, $corrcoef(A,B)=cov(A,B)/(\sigma_A \sigma_B)$, between the two simulations was 0.914. In addition, the root mean squared error, $RMSE(A,B)={[\frac{1}{n} \sum_{i=1}^{n}(A_i-B_i)^2]}^{1/2}$, was only 1.018, i.e., the error is around 1 out of 100 trials. 
These results demonstrated that the MCMC simulation can predict our dynamic simulation traversal result very well. Considering that the MCMC method can finish the 100-trial prediction within 10 seconds whereas the dynamics simulation took around 100 hours, there is a huge advantage in saving computational time by using the MCMC method (by a factor of $10^5$) compared to using the full dynamic simulation.

To evaluate the generality of this result, we tested three other sets of randomly chosen initial states and found excellent match throughout (Table \ref{tab5}).

\begin{table}
\setlength{\abovecaptionskip}{-0.1cm}
\setlength{\belowcaptionskip}{0cm}
\caption{Prediction Evaluation Using Different Initial States.}
\label{tab5}
\begin{center}
\begin{tabular}{ccccc}
\hline
$d$ (m) & $v_x$ (m/s) & $v_y$ (m/s) & $corrcoef$ & $RMSE$   \\ \hline
0     & 0        & 15       & 0.914    & 1.018  \\
23    & $-$11      & 16       & 0.9097   & 1.133  \\
4     & 5        & 10       & 0.8929   & 1.252  \\
$-$17   & 13       & 12       & 0.9026   & 0.9813 \\
\hline
\end{tabular}
\end{center}
\vspace{0cm}
\end{table}

\section{DISCUSSION}
\label{chap4}
In this study, we developed a stochastic dynamic model to simulate a self-propelled, minimalistic locomotor body with random forces traversing beam obstacles in two dimensions on a horizontal plane. We found that a larger forward propulsive force and a sufficient but not excessive lateral random force facilitate traversal, by providing sufficient kinetic energy to overcome potential energy barriers from obstacle interaction. Asymmetry in beam stiffness induces asymmetric distribution of traversal trajectories, with higher probability to traverse on the side with a lower barrier. The model of interaction with a basic unit of obstacles (two beams) was further extended to simulate the traversal of a multi-obstacle field. Finally, we applied a Markov chain Monte Carlo method to model how the body statistically transitions between adjacent obstacle units, which well predicted the probability distribution of the final location of the locomotor in the large obstacle field.

Our results demonstrated that the potential energy landscape is useful for understanding traversal dynamics without solving them exactly. This would be particularly useful for modeling systems with complex interaction, which is common in animal and robot locomotion in complex terrain. In addition, we can redesign the landscape barriers and basins by adjusting the model parameters that define physical interaction, which can be applied to robot motion control and path planning. Furthermore, stochasticity, which is common in locomotion and usually considered a nuisance \cite{othayoth2020energy,othayoth2021locomotor,xuan2020randomness,fu2020robotic}, can be beneficial for robots when they are trapped by obstacles and need to overcome potential energy barriers. Finally, our exploration of multi-obstacle field traversal supported the plausibility to decompose natural terrain as a combination of multiple obstacles \cite{othayoth2021locomotor}. Our results suggested that, by statistically understanding and predicting the interaction with each type of obstacle unit, larger scale traversal processes may be rapidly predicted.

Future work should add more physical realities such as locomotor degrees of freedom and frictional interaction with the ground to the minimalistic model and expand to 3-D obstacle traversal in 3-D terrain. The addition of the ability to sense, control, and plan obstacle interaction \cite{xuan2021environmental} will help advance robotic traversal of cluttered obstacles \cite{othayoth2021locomotor,Chen2022the}. 


\addtolength{\textheight}{-12cm}   





\section*{ACKNOWLEDGMENT}

We thank Noah Cowan, Ratan Othayoth, and Yaqing Wang for discussion and three anonymous reviewers for suggestions.


\bibliographystyle{IEEEtran}
\bibliography{IEEEabrv,reference}

\begin{thebibliography}{10}
\providecommand{\url}[1]{#1}
\csname url@rmstyle\endcsname
\providecommand{\newblock}{\relax}
\providecommand{\bibinfo}[2]{#2}
\providecommand\BIBentrySTDinterwordspacing{\spaceskip=0pt\relax}
\providecommand\BIBentryALTinterwordstretchfactor{4}
\providecommand\BIBentryALTinterwordspacing{\spaceskip=\fontdimen2\font plus
\BIBentryALTinterwordstretchfactor\fontdimen3\font minus
  \fontdimen4\font\relax}
\providecommand\BIBforeignlanguage[2]{{%
\expandafter\ifx\csname l@#1\endcsname\relax
\typeout{** WARNING: IEEEtran.bst: No hyphenation pattern has been}%
\typeout{** loaded for the language `#1'. Using the pattern for}%
\typeout{** the default language instead.}%
\else
\language=\csname l@#1\endcsname
\fi
#2}}

\bibitem{borenstein1991vector}
J.~Borenstein, Y.~Koren, \emph{et~al.}, ``The vector field histogram-fast
  obstacle avoidance for mobile robots,'' \emph{IEEE Transactions on Robotics
  and Automation}, vol.~7, no.~3, pp. 278--288, 1991.

\bibitem{rimon1990exact}
E.~Rimon, ``Exact robot navigation using artificial potential functions,''
  Ph.D. dissertation, Yale University, 1990.

\bibitem{khatib1986real}
O.~Khatib, ``Real-time obstacle avoidance for manipulators and mobile robots,''
  in \emph{Autonomous robot vehicles}.\hskip 1em plus 0.5em minus 0.4em\relax
  Springer, 1986, pp. 396--404.

\bibitem{thrun2010toward}
S.~Thrun, ``Toward robotic cars,'' \emph{Communications of the ACM}, vol.~53,
  no.~4, pp. 99--106, 2010.

\bibitem{harley2009characterization}
C.~Harley, B.~English, and R.~Ritzmann, ``Characterization of obstacle
  negotiation behaviors in the cockroach, \textit{Blaberus discoidalis},''
  \emph{Journal of Experimental Biology}, vol. 212, no.~10, pp. 1463--1476,
  2009.

\bibitem{li2015terradynamically}
C.~Li, A.~O. Pullin, D.~W. Haldane, H.~K. Lam, R.~S. Fearing, and R.~J. Full,
  ``Terradynamically streamlined shapes in animals and robots enhance
  traversability through densely cluttered terrain,'' \emph{Bioinspiration \&
  Biomimetics}, vol.~10, no.~4, p. 046003, 2015.

\bibitem{gart2018body}
S.~W. Gart and C.~Li, ``Body-terrain interaction affects large bump traversal
  of insects and legged robots,'' \emph{Bioinspiration \& Biomimetics},
  vol.~13, no.~2, p. 026005, 2018.

\bibitem{gart2018dynamic}
S.~W. Gart, C.~Yan, R.~Othayoth, Z.~Ren, and C.~Li, ``Dynamic traversal of
  large gaps by insects and legged robots reveals a template,''
  \emph{Bioinspiration \& Biomimetics}, vol.~13, no.~2, p. 026006, 2018.

\bibitem{othayoth2020energy}
R.~Othayoth, G.~Thoms, and C.~Li, ``An energy landscape approach to locomotor
  transitions in complex 3d terrain,'' \emph{Proceedings of the National
  Academy of Sciences}, vol. 117, no.~26, pp. 14\,987--14\,995, 2020.

\bibitem{han2021shape}
Y.~Han, R.~Othayoth, Y.~Wang, C.-C. Hsu, R.~de~la Tijera~Obert, E.~Francois,
  and C.~Li, ``Shape-induced obstacle attraction and repulsion during dynamic
  locomotion,'' \emph{The International Journal of Robotics Research}, vol.~40,
  no. 6-7, pp. 939--955, 2021.

\bibitem{wang2021cockroaches}
Y.~Wang, R.~Othayoth, and C.~Li, ``Cockroaches adjust body and appendages to
  traverse cluttered large obstacles,'' \emph{bioXiv preprint}, 2021, bioXiv:
  2021.10.02.462900.

\bibitem{mi2021omniroach}
J.~Mi, Y.~Wang, and C.~Li, ``Omni-roach: A legged robot capable of traversing
  multiple types of large obstacles and self-righting,'' \emph{arXiv preprint},
  2021, arXiv: 2112.10614 [cs.RO].

\bibitem{xuan2021environmental}
Q.~Xuan, Y.~Wang, and C.~Li, ``Environmental force sensing enables robots to
  traverse cluttered obstacles with interaction,'' \emph{arXiv preprint}, 2021,
  arXiv: 2112.07900 [cs.RO].

\bibitem{othayoth2021locomotor}
R.~Othayoth, Q.~Xuan, Y.~Wang, and C.~Li, ``Locomotor transitions in the
  potential energy landscape-dominated regime,'' \emph{Proceedings of the Royal
  Society B: Biological Sciences}, vol. 288, no. 1949, p. 20202734, 2021.

\bibitem{Chen2022the}
C.~Li and K.~Lewis, ``The need for and feasibility of alternative ground robots
  to traverse sandy and rocky extraterrestrial terrain,'' \emph{Advanced
  Intelligent Systems}, p. 2100195, 2022.

\bibitem{bergman2000caribou}
C.~M. Bergman, J.~A. Schaefer, and S.~Luttich, ``Caribou movement as a
  correlated random walk,'' \emph{Oecologia}, vol. 123, no.~3, pp. 364--374,
  2000.

\bibitem{faisal2008noise}
A.~A. Faisal, L.~P. Selen, and D.~M. Wolpert, ``Noise in the nervous system,''
  \emph{Nature Reviews Neuroscience}, vol.~9, no.~4, pp. 292--303, 2008.

\bibitem{moore2017unpredictability}
T.~Y. Moore, K.~L. Cooper, A.~A. Biewener, and R.~Vasudevan, ``Unpredictability
  of escape trajectory explains predator evasion ability and microhabitat
  preference of desert rodents,'' \emph{Nature Communications}, vol.~8, no.~1,
  pp. 1--9, 2017.

\bibitem{xuan2020randomness}
Q.~Xuan and C.~Li, ``Randomness in appendage coordination facilitates strenuous
  ground self-righting,'' \emph{Bioinspiration \& Biomimetics}, vol.~15, no.~6,
  p. 065004, 2020.

\bibitem{grand2010motion}
C.~Grand, F.~Benamar, and F.~Plumet, ``Motion kinematics analysis of
  wheeled--legged rover over 3d surface with posture adaptation,''
  \emph{Mechanism and Machine Theory}, vol.~45, no.~3, pp. 477--495, 2010.

\bibitem{onuchic2004theory}
J.~N. Onuchic and P.~G. Wolynes, ``Theory of protein folding,'' \emph{Current
  Opinion in Structural Biology}, vol.~14, no.~1, pp. 70--75, 2004.

\bibitem{chodera2006long}
J.~D. Chodera, W.~C. Swope, J.~W. Pitera, and K.~A. Dill, ``Long-time protein
  folding dynamics from short-time molecular dynamics simulations,''
  \emph{Multiscale Modeling \& Simulation}, vol.~5, no.~4, pp. 1214--1226,
  2006.

\bibitem{schmitt2000mechanical}
J.~Schmitt and P.~Holmes, ``Mechanical models for insect locomotion: dynamics
  and stability in the horizontal plane {I}. {Theory},'' \emph{Biological
  Cybernetics}, vol.~83, no.~6, pp. 501--515, 2000.

\bibitem{ames2005sufficient}
A.~D. Ames, A.~Abate, and S.~Sastry, ``Sufficient conditions for the existence
  of {Zeno} behavior,'' \emph{Proceedings of the 44th IEEE Conference on
  Decision and Control}, pp. 696--701, 2005.

\bibitem{gilks1995markov}
W.~R. Gilks, S.~Richardson, and D.~Spiegelhalter, \emph{Markov chain Monte
  Carlo in practice}.\hskip 1em plus 0.5em minus 0.4em\relax CRC press, 1995.

\bibitem{fu2020robotic}
Q.~Fu and C.~Li, ``Robotic modelling of snake traversing large, smooth
  obstacles reveals stability benefits of body compliance,'' \emph{Royal
  Society Open Science}, vol.~7, no.~2, p. 191192, 2020.

\end{thebibliography}

\end{document}